%% file: main.tex
\documentclass[10pt,twocolumn,letterpaper]{article}

\usepackage{cvpr}              

\input{preamble}

\definecolor{cvprblue}{rgb}{0.21,0.49,0.74}
\usepackage[pagebackref,breaklinks,colorlinks,allcolors=cvprblue]{hyperref}

\def\paperID{xxxx} 
\def\confName{CVPR}
\def\confYear{2025}

\title{$\text{T}^2\text{SG}$: Traffic Topology Scene Graph for Topology Reasoning \\ in Autonomous Driving}

\author{Changsheng Lv \qquad Mengshi Qi\thanks{Corresponding author}\qquad Liang Liu\qquad Huadong Ma \vspace{1mm} \\
 State Key Laboratory of Networking and Switching Technology, \\ Beijing University of Posts and Telecommunications, China\\
 {\it \{lvchangsheng, qms, liangliu, mhd\}@bupt.edu.cn} 
}

\begin{document}
\maketitle
\input{sec/0_abstract}    
\input{sec/1_intro}
\input{sec/2_related_work}
\input{sec/3_method}
\input{sec/4_expeiments}
\input{sec/5_conclusion}
\input{sec/6_acknowledgments}
{
    \small
    \bibliographystyle{ieeenat_fullname_etal}
    \bibliography{main}
}


\end{document}

%% file: preamble.tex
\usepackage{algorithm}
\usepackage{algorithmic}
\usepackage{multirow}
\usepackage{booktabs}
\usepackage{colortbl} 
\usepackage{graphicx} 
\usepackage{amsmath} 
\makeatletter
\newcommand{\thickhline}{%
	\noalign {\ifnum 0=`}\fi \hrule height 1pt
	\futurelet \reserved@a \@xhline
}

%% file: sec/0_abstract.tex
\begin{abstract}
Understanding the traffic scenes and then generating high-definition~(HD) maps present significant challenges in autonomous driving. In this paper, we defined a novel {\it \underline{T}raffic \underline{T}opology \underline{S}cene \underline{G}raph}~($\text{T}^2\text{SG}$), a unified scene graph explicitly modeling the lane, controlled and guided by different road signals~({\it e.g.}, right turn), and topology relationships among them, which is always ignored by previous high-definition (HD) mapping methods. For the generation of $\text{T}^2\text{SG}$, we propose TopoFormer, a novel one-stage {\it \underline{Topo}logy Scene Graph Trans\underline{Former}} with two newly-designed layers. Specifically, TopoFormer incorporates a Lane Aggregation Layer~(LAL) that leverages the geometric distance among the centerline of lanes to guide the aggregation of global information. Furthermore, we proposed a Counterfactual Intervention Layer~(CIL) to model the reasonable road structure~({\it e.g.}, intersection, straight) among lanes under counterfactual intervention. Then the generated $\text{T}^2\text{SG}$ can provide a more accurate and explainable description of the topological structure in traffic scenes. Experimental results demonstrate that TopoFormer outperforms existing methods on the $\text{T}^2\text{SG}$ generation task, and the generated $\text{T}^2\text{SG}$ significantly enhances traffic topology reasoning in downstream tasks, achieving a state-of-the-art performance of 46.3 OLS on the OpenLane-V2 benchmark. Our source code is available at https://github.com/MICLAB-BUPT/T2SG.
\end{abstract}

%% file: sec/1_intro.tex
\section{Introduction}
\label{sec:intro}

\begin{figure}[t]
    \centering
    \includegraphics[width=1\linewidth]{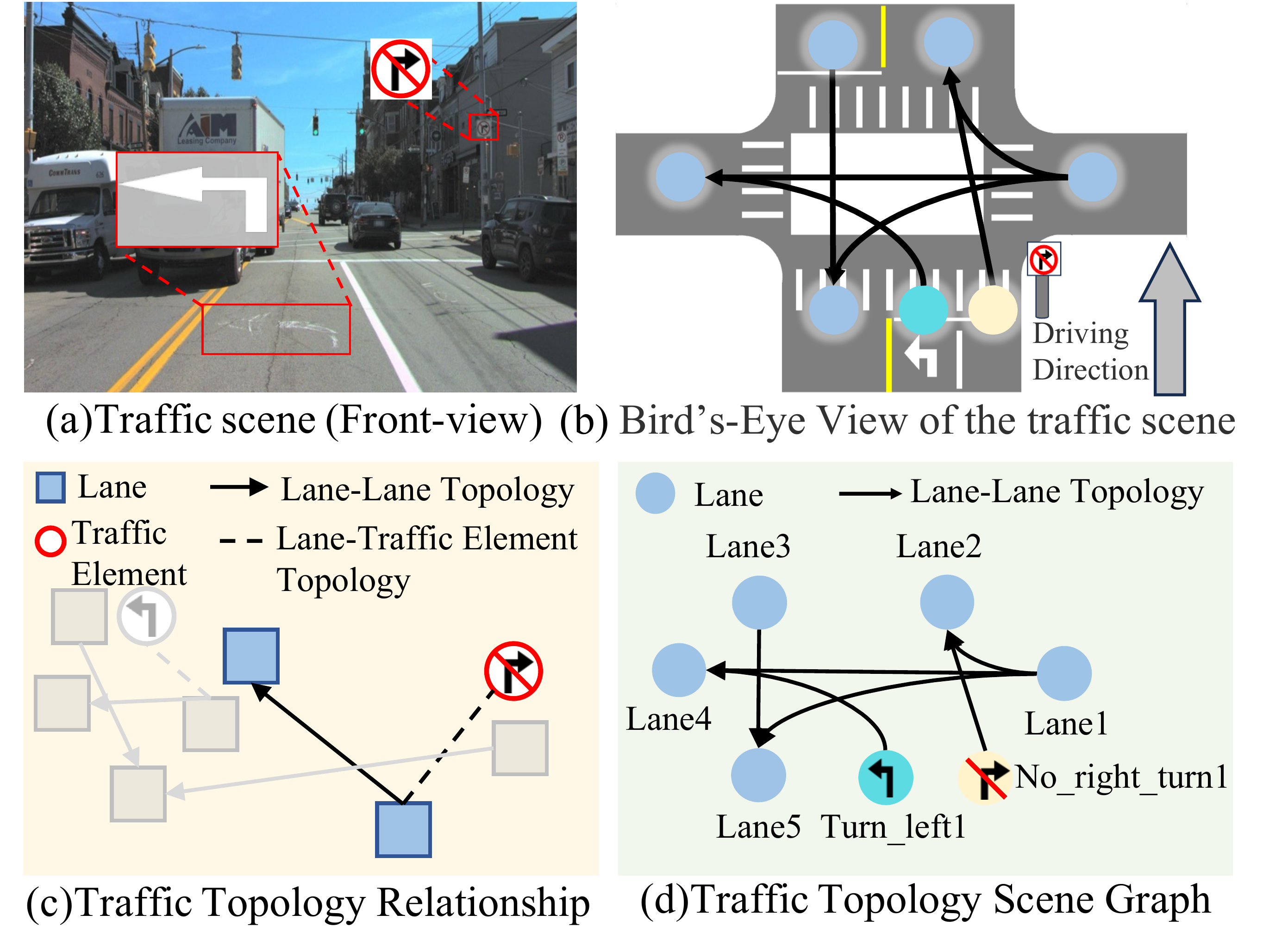}
    \caption{An example of a traffic scene understanding is illustrated as follows: (a) The traffic scene. (b) A BEV of the traffic scene, (c) The topology relationship proposed in TopoNet~\cite{li2023graph}, (d) The $\text{T}^2\text{SG}$ proposed in our work. Different from TopoNet, $\text{T}^2\text{SG}$ can simultaneously model the whole relationships in the scene graph.
    }
    \label{fig: an overview of an example of T2SGG}
\end{figure}

Understanding the traffic scene is the key component of autonomous driving. Except for detecting and recognizing individual elements, the vehicles need to infer the topology relationship among them. Conventional traffic scene understanding tasks, such as lane perception~\cite{tabelini2021keep}, road signal elements detection~\cite{carion2020end}, and high definition~(HD) mapping~\cite{liao2022maptr} focus primarily on the  {\it isolated} elements~({\it i.e.}, map elements, and road signal elements ) but miss the relationship among them. To model the unified HD mapping for lane perception and road signal elements association, the traffic topology reasoning task on OpenLane-V2~\cite{wang2023openlanev2} has recently been proposed. As a map-like reasoning results shown in Figure~\ref{fig: an overview of an example of T2SGG}(c), the reasoning task aims to create a topology graph of the detected elements and thus facilitate decision-making in the downstream tasks, such as ego planning~\cite{casas2021mp3} and motion prediction~\cite{wang2024drones,zhu2023unsupervised,qi2019stagnet}.

A primary challenge in traffic topology reasoning involves accurately modeling intricate traffic scene structures from multi-view camera inputs. Existing HD mapping methods~\cite{liu2023vectormapnet} explicitly model the spatial relation between lanes but overlook the control and guide relationship between road signal elements and lanes. 
TopoNet~\cite{li2023graph} has noticed this issue and proposed a graph-based method that treats the lanes and road signal elements as nodes and constructed a heterogeneous typology graph to describe the aforementioned relationships. However, this method neglects the control and guide information inherent in the traffic rules represented by each road signal. As illustrated in Fig~\ref{fig: an overview of an example of T2SGG} (b), a lane guided by the ``Turn left" signal will only establish a connection with the lane left. Similarly, the ``No right turn'' signal also carries corresponding semantics.

To fully leverage this control and guide information to enhance lane centerline detection and traffic topology reasoning. Inspired by Scene Graph Generation~\cite{xu2017scene}, we explicitly model lanes guided by different road signals and their topology relationship using a unified traffic scene graph. In contrast to such scene graphs~\cite{xu2017scene}, the major challenge of the unified traffic scene graph we proposed is how to reason about relationships by simultaneously considering the spatial information of lanes and corresponding control and guide information of road signals. Compared to the independent traffic-lane and lane-lane topology reasoning tasks defined in OpenLane-V2~\cite{wang2023openlanev2}, the unified traffic scene graph facilitates joint learning of these two tasks.

Based on the above analysis, as illustrated in Fig~\ref{fig: an overview of an example of T2SGG} (d), we define a novel {\it \underline{T}raffic \underline{T}opology \underline{S}cene \underline{G}raph}~($\text{T}^2\text{SG}$), whose goal is to generate a visually-grounded scene graph from the input multi-view images. In the $\text{T}^2\text{SG}$, an object instance is characterized by a centerline with a corresponding category label and a relationship is characterized by a directed edge between two centerlines with a binary value \{0,1\} indicating whether they are connected. To solve the complex relation reasoning problem, we propose a one-stage traffic topology scene graph generator {\bf TopoFormer}, which stands for {\it \underline{Topo}logy scene graph Trans\underline{Former}}. TopoFormer contains a Lane Aggregation Layer that aggregates features according to the spatial proximity of the lane via geometry-guided self-attention. In this way, TopoFormer effectively obtains lane embeddings with the global context.

Furthermore, we capture the reasonable road structure among lanes in the traffic scene via the Counterfactual Intervention Layer (CIL), encompassing simple structures such as straight roads and more complex structures like crossroads and multi-way intersections. Previous methods~\cite{liu2023vectormapnet, fu2024topologic} focus on utilizing the spatial positions of centerlines to make {\it local} predictions of the lane relationships. However, these methods ignore the reasonable road structure in the real traffic scene, whereas joint reasoning with road structure can often resolve ambiguous relationships that arise from {\it local} predictions in isolation. Specifically, we consider the self-attention weights among lanes to signify the road structure and compare the factual structure ({\it i.e.}, the learned attention weights) with the counterfactual structure ({\it i.e.}, zero attention weights) on the ultimate prediction ({\it i.e.}, the output score). The proposed CIL can enhance the learned road structure's total indirect effect (TIE) on the prediction results. 

Our main contributions can be summarized as follows:

\par\textbf{(1)}~We propose the first unified Traffic Topology Scene Graph~($\text{T}^2\text{SG}$) to explicitly model the lanes, which are controlled and guided by different road signals, and the topology relationships among these lanes. 

\par\textbf{(2)}~We propose a Topology Scene Graph Transformer~(TopoFormer) for $\text{T}^2\text{SG}$ task, which captures global dependencies among lanes with a Lane Aggeration Layer. 

\par\textbf{(3)}~We introduce a Counterfactual Intervention Layer to emphasize the reasonable road structure influencing lane connectivity and effectiveness in topology reasoning tasks.

\par\textbf{(4)}~We evaluate our TopoFormer in scene graph generation task and show it outperforms all state-of-the-art methods. Furthermore, we attain a 46.3 OLS on the traffic topology reasoning benchmark, OpenLane-V2~\cite{wang2023openlanev2}, demonstrating the effectiveness of the proposed $\text{T}^2\text{SG}$ and TopoFormer for downstream tasks.

%% file: sec/2_related_work.tex
\section{Related Work}
\label{sec:formatting}

\noindent{\bf Scene Graph Generation.}
Scene graphs were first introduced for Image Retrieval~\cite{johnson2015image,qi2021semantics}, which broke down an image into its constituent objects, their attributes, and the relationships between them. The Visual Genome dataset~\cite{krishna2017visual} advanced image understanding, enabling scene graph extraction methods~\cite{cong2023reltr,im2024egtr,qi2020stc,qi2019attentive}. Scene graphs are increasingly applied in tasks like image captioning~\cite{yang2024exploring} and visual question answering~\cite{wang2023vqa}. \cite{wald2020learning} introduced 3D scene graphs for object relationship perception, with subsequent research exploring GCN and Transformer-based methods~\cite{lv2024sgformer} for point cloud registration~\cite{sarkar2023sgaligner} and scene generation~\cite{zhai2024commonscenes}. Unlike prior approaches, $\text{T}^2\text{SG}$ pioneers scene graphs in traffic scene understanding, modeling lanes as nodes and their connections as edges.

\begin{figure*}[!t]
    \centering
    \includegraphics[width=0.95\linewidth]{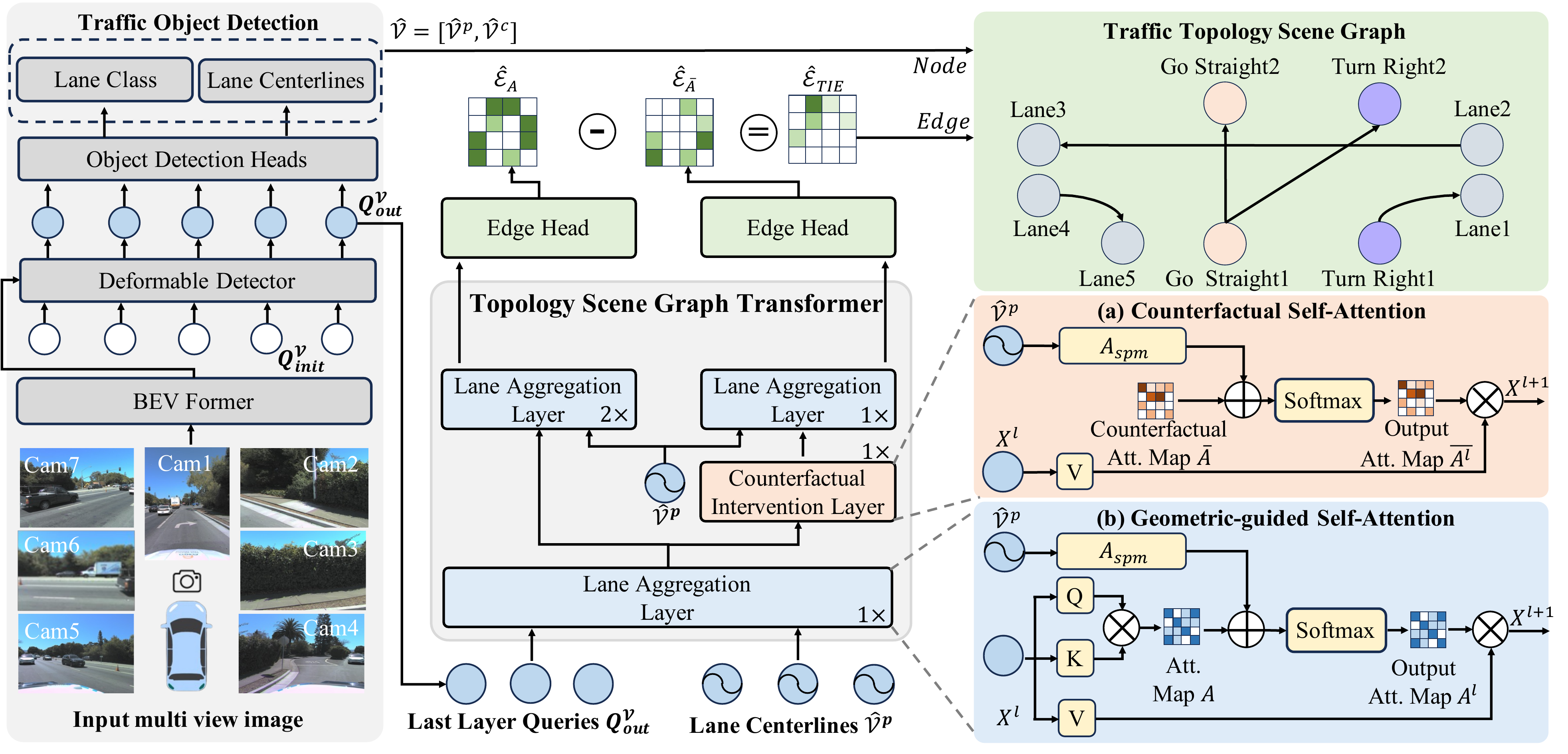}
    \caption{{\bf The overview of our proposed TopoFormer.} Given the input multi-view images, we employ a DETR-like detector to identify lane objects with corresponding class and centerline coordinates. Subsequently, TopoFormer infers the relationships among these objects, which, along with the objects themselves, constitute the $\text{T}^2\text{SG}$. The main components of TopoFormer include two newly designed layers: (a) the  Counterfactual Intervention Layer incorporating Counterfactual Self-Attention, and (b) the Lane Aggregation Layer incorporating Geometric-guided Self-Attention. Ultimately, the output of TopoFormer is a traffic topology scene graph, encapsulating the topological relationships among lanes, and guided by various road signals associated with the lanes.}
    \label{fig: Main model}
\end{figure*}

\noindent{\bf Traffic Topology Reasoning.}
STSU~\cite{can2021structured} introduced a lane topology reasoning approach for road structure comprehension, focusing on Bird's Eye View (BEV) lane centerline detection through three stages: BEV feature construction, centerline detection, and connection prediction. TPLR~\cite{can2022topology} and LaneGAP~\cite{liao2025lane} refined centerline representation, enhancing continuity and shape accuracy. Online map construction methods like MapTR~\cite{liao2022maptr}, VectorMapNet~\cite{liu2023vectormapnet}, BeMapNet~\cite{qiao2023end}, and Gemap~\cite{zhang2024online} integrated lane topology reasoning with geometric information, modeling map elements such as lane lines, pedestrian crossings, and curbs. TopoNet addressed complex road scenarios by using Graph Neural Networks to connect driving lanes and traffic signs. TopoSeq~\cite{yang2025topo2seq} proposed randomized prompt-to-sequence learning for joint extraction of lane topology from Directed Acyclic Graphs and geometric lane graphs. TopoMLP~\cite{wutopomlp} and TopoLogits~\cite{fu2024topologic} enhanced topology inference by leveraging lane spatial positions. In contrast, our method employs a geometric-guided Lane Aggregation Layer, introducing spatial information to better illustrate lane relationships rather than augmenting lane features.

\noindent{\bf Counterfactual Intervention.}
Counterfactual Intervention (CI) is widely used in reasoning tasks like VQA~\cite{lv2024disentangled}, Re-ID~\cite{rao2021counterfactual}, and scene understanding~\cite{huang2024causalpc}. ~\cite{lv2024disentangled} modeled the physical knowledge relationships among different objects and apply them as counterfactual interventions to boost causal reasoning, while ~\cite{rao2021counterfactual} enhanced Re-ID by applying CI to feature maps, maximizing task-relevant attention. Inspired by these, we leverage lane geometry to construct road structures and apply CI, maximizing the Total Indirect Effect (TIE) to encourage the model to learn more reasonable road structures for topology reasoning.

%% file: sec/3_method.tex
\section{Proposed Approach}
\subsection{Overview}
\noindent{\bf Problem Definition.}
We define a traffic topology scene graph, denoted as $\text{T}^2\text{SG}$, as \( \mathcal{G} = (\mathcal{V}, \mathcal{E}) \). This graph represents the traffic scene, detailing each lane's centerline coordinates, category, and connectivity with other lanes. Each lane $v_i \in \mathcal{V}$ has a lane category label $v_i^c$ from a set of lane categories $\mathcal{C}_{lc}$, and an ordered list of 3D points $v_i^p$, which are sampled from the centerline along the direction of the lane at a fixed frequency ({\it i.e.}, 2Hz) following~\cite{wang2023openlanev2}, denoted as $v_i=[v_i^c, v_i^p]$. Specifically, for a centerline $v_i^p=[p_1, p_2, ..., p_{l}]$, $p_1=(x_1, y_1, z_1)$ represents the starting point of the lane, while $p_l=(x_l, y_l, z_l)$ denotes the ending point.  The direction of a centerline, from the starting point to the ending point, denotes that a vehicle in this lane should follow the direction. Each edge $e_{ij} \in \mathcal{E}$ represents the connectivity relationship between lane $v_i$ and lane $v_j$, where $i \neq j$ ensures that a lane does not connect to itself. Since a lane centerline is directed and represented as a list of points, the connection of two lanes means that the ending point of a centerline $p_1$ is connected to the starting point of another centerline $p_l$.

As illustrated in Figure~\ref{fig: Main model}, based on a DETR-liked~\cite{carion2020end} lane centerlines detector, the overall framework of our proposed method follows typical Transformer architecture but consists of two carefully designed components: {\it Lane Aggregation Layer}~({\bf LAL}) captures global dependencies among lanes with geometry-guided self-attention; {\it Counterfactual Intervention Layer}~({\bf CIL}) capture the reasonable road structure through counterfactual self-attention. For the edge~$\mathcal{E}$, we use the total indirect effect $\hat{\mathcal{E}}_{\text{TIE}}$ calculated during the training stage as the output, while only using the normal predictions $\hat{\mathcal{E}}_A$ ($A$ represents utilizing attention weights without counterfactual intervention.) in the inference stage. This will be detailed in the Training and Inference section.

\noindent{\bf Lane Centerline Detection.}
Given multi-view images $\mathcal{I}= \left\{ I_i \in \mathbb{R}^{3 \times H_I \times W_I }\mid i=1, 2, ..., N_I \right\}$ from $N_I$ multi-camera views, where $H_I$ and $W_I$ represent the height and width of the input images, respectively. The backbone networks such as ResNet-50~\cite{he2016deep} and FPN~\cite{lin2017feature} are utilized to extract multi-view 2D features $\mathcal{F}_{2D}$. Based on 2D features $\mathcal{F}_{2D}$, we employ a simplified view Transformer, as proposed in BEVformer~\cite{li2022bevformer}, to generate grid BEV feature $\mathcal{F}_{BEV}$, and use it as the input for the Deformable Detector~\cite{zhudeformable}, represented as:
\begin{equation}
\begin{aligned}
&Q_{out}^{\mathcal{V}} = \text{DeformDETR}(Q_{init}^{\mathcal{V}}, \mathcal{F}_{BEV}), \\
&\hat{\mathcal{V}}^{p}, \hat{\mathcal{V}}^{c} = \text{Lane Head}(Q_{out}),
\end{aligned}
\label{eq:lc_det}
\end{equation}
where $Q_{init}^{\mathcal{V}}, Q_{out}^{\mathcal{V}}\in\mathbb{R}^{N\times256}$ denote the initialized query and the output query from the final layer, respectively, and $N$ signifies the number of queries. \(\hat{\mathcal{V}} = [\hat{\mathcal{V}}^{p}, \hat{\mathcal{V}}^{c}]\), where \(\hat{\mathcal{V}}^{p} = \left\{ \hat{v}_i^{p} \right\}_{i=1}^{N}\) and \(\hat{\mathcal{V}}^{c} = \left\{ \hat{v}_i^{c} \right\}_{i=1}^{N}\) denote the predicted nodes in the graph \(\hat{\mathcal{G}}\), and   $\left\{ \hat{{v}}_i^{p} \in \mathbb{R}^{l \times 3}, \hat{{v}}_i^{c} \in {C}_{lc} \mid i=1,2,.., N\right\}$ signifies the ordered points list of the lane centerlines and class of the lane, respectively. The Lane Head($\cdot$) is constructed by two independent multilayer perceptron (MLPs) to predict the points of centerlines and the classification scores of lanes.

\subsection{Lane Aggregation Layer }
The Lane Aggregation Layer (LAL) is designed to leverage the geometric distance among the centerline of lanes to guide the aggregation of global structural information, which is a Transformer-encoder-like layer with the core component of Geometry-guided Self-Attention (GSA). Following~\cite{li2023graph}, we utilize the output query $Q_{out}^{\mathcal{V}}$ as the input lane feature and predicted centerlines $\hat{\mathcal{V}}^{p}$ as the geometric information for the lane centerline. These are then fed into the Lane Aggregation Layer. The input lane features are passed through linear projections to be embedded into a $d$-dimensional hidden feature $X^0 \in \mathbb{R}^{N \times d}$. The output $X^{l}$ is the feature encoded by $l$ layers of LAL.

\noindent{\bf Geometry-guided Self-Attention~(GSA)} is proposed in the layer for the message passing in the graph, which is different from the conventional self-attention, as shown in Figure~\ref{fig: Main model}. Inspired by~\cite{fu2024topologic}, the geometric distance among lane centerlines can serve as the basis for global dependencies, thereby enhancing the accuracy of lane topology reasoning. Therefore, we introduce the spatial proximity matrix~(SPM)~\cite{zhao20213dvg} to describe the normalized inverse geometric distances among lanes. It can be formulated as:
\begin{equation}
    A_{SPM}=\text{Norm}\left(\frac{1}{d(\hat{v}_{i,l}^{p}, \hat{v}_{j,0}^{p}) + \epsilon}\right), i,j \in N,
\end{equation}
where $\hat{v}_{i,l}^{p}$ is the end point of predicted centerline $\hat{v}_i^p$, and $\hat{v}_{j,0}^{p}$ is the start point of predicted $\hat{v}_j^p$. $\epsilon$ is a small constant to avoid infinity, $d(\cdot)$ denotes the distance ({\it i.e.}, ${\ell_1}$ distance), and $\text{Norm}$ is a normalization operation that divides each entry in the $A_{SPM}$ by the mean inverse distance. 
Our core idea is to aggregate the global information of the lanes based on their spatial distances. Therefore, as shown in Figure~\ref{fig: Main model}(b), we add $A_{SPM}$ to the self-attention, the GSA can be formulated as the follows in the layer $l$:
\begin{equation}
    \text{GSA}(X^l)= A^l \cdot X^lW_V^l
\end{equation}
where
\begin{equation}
    A^l=\text{softmax}\left(\frac{X^lW_Q^l \cdot({X^lW_K^l})^{\top}}{\sqrt{d}}+ A_{SPM} \right), \label{eq: LAL_2}
\end{equation}
where $W_Q^l, W_K^l, W_V^l \in \mathbb{R}^{d \times d}$ are the weights of linear layer, $\cdot$ denotes the matrix multiplication, $A^l \in \mathbb{R}^{N \times N}$ denotes the attention weights in the $l^{th}$ lane aggeration layer, $d$ and $N$ denoted the hidden dimension and number of queries. Subsequently, the lane feature $X^{l+1}$ passed into the feed-forward network (FFN) preceded and succeeded by residual connections and normalization layers, and the output of $(l+1)$-th LAL can be formulated as:
\begin{equation}
    X^{l+1} =\text{Norm}(X^l+\text{FFN}(\text{GSA}(X^l))),
\label{GSA:overall}
\end{equation}
It is noteworthy that different from TopoMLP~\cite{wutopomlp} which introduces position encoding to enhance the features of each lane centerline, our proposed LAL explicitly utilizes position encoding for the aggregation of global information (\emph{i.e.,} spatial interaction among lanes), to exhibit stronger generalization.

\subsection{Counterfactual Intervention Layer}
The Counterfactual Intervention Layer~(CIL) is designed to capture the reasonable road structure among lanes in the traffic scene which is a transformer-encoder-like layer with the core component of Counterfactual Self-Attention (CSA).
According to the analysis by \cite{fu2024topologic}, geometry-based aggregation heavily relies on the detected lane centerlines \(\hat{\mathcal{V}}^{p}\) in Eq.(\ref{eq:lc_det}). Inaccuracies in centerline detection can interfere with the quality of the $A_{SPM}$ and lead to erroneous relationship predictions. We propose to leverage lane feature self-attention weights to represent the learned road structure and utilize counterfactual road structures ({\it e.g.}, zero attention) to improve the learning of traffic scene structure. 

\noindent{\bf Counterfactual Self-Attention} is designed within the layer to predict relationships among lanes. Drawing inspiration from causal inference methodologies~\cite{rao2021counterfactual}, we propose a counterfactual intervention to explore the effects of the learned attention weights. Specifically, we perform the counterfactual intervention $do(A=\overline{A})$ by creating a hypothetical attention weight $\overline{A}$ to replace the original, while maintaining the lane feature $X$ and $A_{SPM}$ unchanged. The structure of the counterfactual self-attention closely resembles that of the geometry-guided self-attention. However, the primary distinction lies in the configuration of the attention weights:
\begin{equation}
    \text{CSA}(X^l)=\text{softmax}\left( \overline{A^l} + A_{SPM}\right) \cdot X^lW_V^l,
\end{equation}
where 
\begin{equation}
    \overline{A^l}=\text{Zeros}\left(\frac{X^lW_Q^l \cdot({X^lW_K^l})^{\top}}{\sqrt{d}}\right), \label{eq:C_attention}
\end{equation}
where $W_Q^l, W_K^l, W_V^l \in \mathbb{R}^{d \times d}$ are the weights of linear layer, $\text{Zeros}$ denotes the operation of generating a zero matrix with the same shape as the original matrix. $\overline{A^l}$ represents the hypothetical attention weight at the $l^{th}$ layer. The output of $(l+1)$-th CIL can be formulated as:
\begin{equation}
    X^{l+1} =\text{Norm}(X^l+\text{FFN}(\text{CSA}(X^l))),
\end{equation}
where $X^l$ denotes the output of the $l$-th LAL, and the FFN has the same structure but operates independently, as described in Eq.~\ref{GSA:overall}.

\subsection{Edge Prediction Head}
In the graph \(\mathcal{G}(\mathcal{V}, \mathcal{E})\), \(\mathcal{E}\) means the set of relationships or edges between all node pairs \(\left\{[v_i, v_j] \mid v_i, v_j\in \mathcal{V},i \neq j\right\}\), where each edge is represented as a binary value \{0, 1\} indicating whether there is a connection from \(v_i\) to \(v_j\). To obtain the predicted \(\hat{\mathcal{E}}\), we developed an edge prediction head. Specifically, given the final output lane feature $\widetilde{X}$ from our TopoFormer, the edge prediction head first applies two independent three-layer MLPs for the start lane and end lane:
\begin{equation}
\begin{aligned}
    &\widetilde{X}^{\prime}_{s} = \text{MLP}_{s}(\widetilde{X}),
    &\widetilde{X}^{\prime}_{e} = \text{MLP}_{e}(\widetilde{X}),
\end{aligned}
\end{equation}
where the subscript $s$ and $e$ represent the start lane and the end lane, respectively. $\text{MLP}_{s}$ and $\text{MLP}_{e}$ respectively denote the corresponding MLPs for lanes. For each pair of lane $\widetilde{x}_s^{\prime} \in \widetilde{X}^{\prime}_{s}$ and $\widetilde{x}_e^{\prime} \in \widetilde{X}^{\prime}_{e}$, the confidence in their relationship is computed as follows:
\begin{equation}
   \hat{\mathcal{E}}_{s,e} = \text{sigmoid} \Big( \text{MLP}_{\text{edge}} \big( \text{concat}(\widetilde{x_s}^{\prime}, \widetilde{x_e}^{\prime}) \big) \Big), s \neq e
\end{equation}
where the ``concat'' operation combines the feature dimensions and the output dimension of $\text{MLP}_{\text{edge}}$ is 1. The sigmoid function constrains this output to the range [0,1]. 

\subsection{Training and Inference}
During the training phase, our loss is divided into two parts, including the loss of node detection $\mathcal{L}_{\mathcal{V}}$ and the loss of edge prediction $\mathcal{L}_{\mathcal{E}}$. The detection loss for node $\mathcal{L}_{\mathcal{V}}$ is decomposed into a lane category~$\mathcal{V}^{c}$ classification and a centerline~$\mathcal{V}^{p}$ regression loss:
\begin{equation}
    \mathcal{L}_{\mathcal{V}}=\lambda_{cls} \cdot \mathcal{L}_{cls} + \lambda_{reg} \cdot \mathcal{L}_{reg},
    \label{eq: classification}
\end{equation}

\noindent where $\lambda_{cls}$ and $\lambda_{reg}$ are the coefficients.  The classification loss $\mathcal{L}_{cls}$ is a Focal loss~\cite{lin2017focal} and the regression loss $\lambda_{reg}$ is an L1 loss. Note that the regression loss is calculated on the normalized 3D coordinates of the predicted ordered points list $\hat{\mathcal{V}}^{p}$. 
For the loss of edge prediction $\mathcal{L}_{\mathcal{E}}$, the total indirect effect of learned road structure on the prediction can be represented by the difference between the observed prediction $\hat{\mathcal{E}}_{A} = \hat{\mathcal{E}}(do(A=A), X=\widetilde{X})$ and its counterfactual result $\hat{\mathcal{E}}_{\overline{A}}=\hat{\mathcal{E}}(do(A=\overline{A}), X=\widetilde{X})$:
\begin{equation}
    \hat{\mathcal{E}}_{\text{TIE}}=\mathbb{E}_{\overline{A}\sim\gamma}[\hat{\mathcal{E}}_{A} - \hat{\mathcal{E}}_{\overline{A}}],
\end{equation}
where $\widetilde{X}$ denotes the final output lane feature, $\hat{\mathcal{E}}_{\text{TIE}}$ refers to the total indirect effect on the prediction, and  $\gamma$ is the distribution of the counterfactual attentions. Furthermore, we can adapt the focal loss on the $\hat{\mathcal{E}}_{\text{TIE}}$ with the coefficient $\lambda_{cls}$:
\begin{equation}
    \mathcal{L}_{\mathcal{E}}= \lambda_{cls} \cdot \mathcal{L}_\text{cls}(\hat{\mathcal{E}}_{\text{TIE}}, \mathcal{E}_{\text{GT}})
\end{equation}
where $\mathcal{E}_{\text{GT}}$ denotes the ground truth of the edge and the classification loss $\mathcal{L}_{cls}$ is similar to Eq.~(\ref{eq: classification}).

Finally, our loss is the sum of the above losses:
\begin{equation}
    \mathcal{L}_{total} = \mathcal{L}_{\mathcal{V}} + \mathcal{L}_{\mathcal{E}}.
\end{equation}
During the inference phase, we no longer use $\hat{\mathcal{E}}_{\text{TIE}}$ as the predicted result for edges. Instead, we utilize $\hat{\mathcal{E}}_A$, which has not been subjected to counterfactual intervention.

\subsection{Traffic topology reasoning}
To validate the efficacy of our proposed $\text{T}^2\text{SG}$ for downstream tasks, we selected the traffic topology reasoning task introduced by~\cite{wang2023openlanev2} for evaluation. The topology reasoning task involves detecting lane centerlines from multi-view images, 2D traffic elements from the front view image, and discerning relationships among them. The output of this task comprises two components: detection results and topology reasoning results. For the traffic elements detection, similar to the lane detection in Eq.(\ref{eq:lc_det}), we get the multi-level 2D features $\mathcal{F}_{FV}$ obtained from the front view image processed through ResNet-50~\cite{he2016deep} and FPN~\cite{lin2017feature}. These features serve as the input for the Deformable Detector, represented as:
\begin{equation}
\begin{aligned}
&Q_{out}^{T} = \text{DeformDETR}(Q_{init}^T, \mathcal{F}_{FV}), \\
&\hat{T}^{b}, \hat{T}^{c} = \text{TE Head}(Q_{out}^{T}),
\end{aligned}
\label{eq:te_det}
\end{equation}

\noindent where $Q_{out}^{T}, Q_{init}^{T} \in \mathbb{R}^{N_{te} \times 256}$ represent the randomly initialized query and the output query from the last layer, respectively, where $N_{te}$ denotes the number of queries. The elements $\hat{T}^{b}=[\hat{t}_i^{b} \in \mathbb{R}^{4} \mid i=1,2,..., N_{te}]$ and $\hat{T}^{c}=[\hat{t}_i^{c} \in \mathcal{C}_{te} \mid i=1,2,..., N_{te}]$ correspond to the bounding box and class of the traffic elements, respectively. The $\text{TE Head}$ is constructed by multilayer perceptron (MLPs). For lane detection, we employ the detection method of \( \text{T}^2\text{SG} \). 

The topology relationship results include the assignment of traffic elements to lane centerlines, denoted as $E_{lt} \in \mathbb{R}^{N \times N_{te}}$, and the connective relationships $E_{ll} \in \mathbb{R}^{N \times N}$ among lanes. Specifically, for $E_{lt}$, traffic elements and lanes that share the same class indicate connectivity. This connectivity is expressed as follows:
\begin{equation}
  E_{lt}(i, j) =
\begin{cases}
1, & \text{if}~\hat{v}_i^{c}=\hat{t}_j^{c} \\

0, & \text{otherwise.}
\end{cases}
\label{eq:E_{lt}}
\end{equation}
where \(\hat{v}_i^{c}\) and \(\hat{t}_j^{c}\) represent the categories of the lane \(\hat{v}_i\) and traffic element \(\hat{t}_j\), respectively, as obtained from Eq.~(\ref{eq:lc_det}) and Eq.~(\ref{eq:te_det}). It is important to note that since our generated \( \text{T}^2\text{SG} \) focuses solely on the road itself, the category \(\mathcal{C}_{lc}\) includes road signals, whereas \(\mathcal{C}_{te}\) additionally encompasses traffic lights (i.e., red, green, and yellow). For the three types of traffic lights, we use two MLP layers to reduce the feature dimension for each lane and traffic light instance, and then the concatenated feature is sent into another MLP with a sigmoid activation to predict their relationship. For $E_{ll}$, we directly use the results of the edge $\hat{\mathcal{E}}$ from our $\text{T}^2\text{SG}$ model as $E_{ll}$, denoted as $[E_{ll}(i,j)=\hat{e}_{ij} \mid \hat{e}_{ij} \in \hat{\mathcal{E}}, i \neq j]$.

%% file: sec/4_expeiments.tex
\section{Experiments}
\begin{table*}[]
\resizebox{\linewidth}{!}
{
\centering
\begin{tabular}{l|ccccccccc}
\hline\thickhline  \noalign{\smallskip} 
\multicolumn{1}{c|}{\multirow{2}{*}{Method}} & \multicolumn{6}{c}{Node} & \multicolumn{3}{c}{Edge} \\ \cmidrule(r){2-7}  \cmidrule(r){8-10}
\multicolumn{1}{c|}{}       & $\text{AP}_{1.0}$    & $\text{AP}_{2.0}$       & $\text{AP}_{3.0}$   & $\text{mAP}_{1.0}$    & $\text{mAP}_{2.0}$       & $\text{mAP}_{3.0}$   & ${\text{A@1}}_{1.0}$  & ${\text{A@1}}_{2.0}$ & $ {\text{A@1}}_{3.0}$  \\ \hline  \noalign{\smallskip}
Baseline                                    &    10.4        &      34.5               &    53.1        &    4.1   &  10.3     &   14.6     &   8.0   &23.2  & 32.4\\
w/ 3DSSG~\cite{wald2019rio}                                      &  10.7(\textcolor{red}{+0.3})          &  36.1(\textcolor{red}{+1.6})                   &    54.2(\textcolor{red}{+1.1})        &   4.4(\textcolor{red}{+0.3})     &    10.3(\textcolor{red}{+0.0})    &    15.1(\textcolor{red}{+0.5})    &    0.4(\textcolor{blue}{-7.6}) & 2.0(\textcolor{blue}{-21.2})  & 4.7(\textcolor{blue}{-27.7})  \\
w/ EdgeGCN~\cite{zhang2021exploiting}                                  &    10.8(\textcolor{red}{+0.4})        &  36.6(\textcolor{red}{+2.1})                   &     \underline{54.5}(\textcolor{red}{+1.4})     &4.5(\textcolor{red}{+0.4})  &10.3(\textcolor{red}{+0.0})&15.8(\textcolor{red}{+1.2}) &   0.4(\textcolor{blue}{-7.6})    &    2.1(\textcolor{blue}{-21.1})    &   5.4(\textcolor{blue}{-27.0})            \\
w/ EGTR~\cite{im2024egtr}                           & 10.8(\textcolor{red}{+0.4})      &    36.3(\textcolor{red}{+1.8})        &     \underline{54.5}(\textcolor{red}{+1.4})                &    4.5(\textcolor{red}{+0.4})        &   10.3(\textcolor{red}{+0.0})    &   16.0(\textcolor{red}{+1.4})     &   8.1(\textcolor{red}{+0.1})     &   23.6(\textcolor{red}{+0.4}) & 33.5(\textcolor{red}{+1.1})    \\
w/ SGformer~\cite{lv2024sgformer}                                  &   \underline{11.2}(\textcolor{red}{+0.8})         &  \underline{36.8}(\textcolor{red}{+2.3})     &     {\bf 54.8(\textcolor{red}{+1.7})}  &   \underline{4.6}(\textcolor{red}{+0.5})   &   \underline{10.9}(\textcolor{red}{+0.6})     &    \underline{16.5}(\textcolor{red}{+1.9})    & \underline{8.5}(\textcolor{red}{+0.5})  &\underline{24.8}(\textcolor{red}{+1.6}) &\underline{34.6}(\textcolor{red}{+2.2})    \\ \noalign{\smallskip} \hline \noalign{\smallskip}
w/ TopoFormer~(Ours)                                &    {\bf 11.8(\textcolor{red}{+1.4})}        &  {\bf 37.5(\textcolor{red}{+3.0})}          &  {\bf 54.8(\textcolor{red}{+1.7})}              &  {\bf 4.8(\textcolor{red}{+0.7})}   &  {\bf 11.3(\textcolor{red}{+1.0})}     &  {\bf 16.7(\textcolor{red}{+2.1})}   & {\bf 8.8(\textcolor{red}{+0.8})} & {\bf 25.6(\textcolor{red}{+2.4})} &  {\bf 35.6(\textcolor{red}{+3.2})}      \\  \noalign{\smallskip} \hline\thickhline
\end{tabular}
}
\caption{{\bf Comparisons of our model and existing state-of-the-art scene graph generation methods on OpenLane-V2}~\cite{wang2023openlanev2}. The subscripts represent Fréchet distance thresholds in the set of \{1.0, 2.0, 3.0\}. More details are described in Metrics. The best performances are highlighted in {\bf bold}, while the second one is \underline{underlined}. \textcolor{red}{Red} indicates the absolute improvements compared with the baseline, while \textcolor{blue}{blue} indicates the decreases compared with the baseline.}
\label{table: T2SG}
\end{table*}

\begin{table*}[ht]
\centering
\renewcommand{\arraystretch}{1} 
\begin{tabular}{ll|lccccc}
\hline\thickhline \noalign{\smallskip}
{Dataset} & {Method} & {Conference}  & {DET$_l$ $\uparrow$} & {DET$_t$ $\uparrow$} & {TOP$_{ll}$ $\uparrow$} & {TOP$_{lt}$ $\uparrow$} & {OLS$\uparrow$}  \\ \noalign{\smallskip} \hline \noalign{\smallskip}
\multirow{8}{*}{ $subset_A$} & STSU~\cite{can2021structured}                       & ICCV2021                        & 12.7   & 43.0     & 2.9     & 19.8    & 29.3 \\
                            & VectorMapNet~\cite{liu2023vectormapnet}               & ICML2023                        & 11.1   & 41.7   & 2.7     & 9.2     & 24.9 \\
                            & MapTR~\cite{liao2022maptr}                      & ICLR2023                        & 17.7   & 43.5   & 5.9     & 15.1    & 31.0   \\
                            & TopoNet~\cite{li2023graph}                    & Arxiv2023                       & 28.5   & 48.1   & 10.9    & 23.8    & 39.8 \\
                            & TopoMLP~\cite{wutopomlp}                    & ICLR2024                        & 28.3   & {\bf49.5}   & 21.6    & \underline{26.9}    & \underline{44.1} \\
                            & TopoLogic~\cite{fu2024topologic}                    & NeurIPS2024                        & \underline{29.9}   & {47.2}   & {\underline{23.9}}    & 25.4    & \underline{44.1} \\
                            & TopoFormer(Ours)                       & -                               & \bf 34.7(\textcolor{red}{+4.8})   & {\underline{48.2}}   & {\bf 24.1(\textcolor{red}{+0.2})}    & \bf 29.5(\textcolor{red}{+3.6})    & \bf 46.3(\textcolor{red}{+2.2}) \\   \bottomrule \noalign{\smallskip}
                           
\multirow{6}{*}{ $subset_B$}  & STSU~\cite{can2021structured}                       & ICCV2021                        & 8.2    & 43.9   & -       & -       & -    \\
                            & VectorMapNet~\cite{liu2023vectormapnet}          & ICML2023                        & 3.5    & 49.1   & -       & -       & -    \\
                            & MapTR~\cite{liao2022maptr}                       & ICLR2023                        & 15.2   & 54.0     & -       & -  & -    \\
                            & TopoNet~\cite{li2023graph}                    & Arxiv2023                       & 24.3   & \underline{55.0}     & 6.7     & 16.7    & 36.8 \\
                            & TopoLogic~\cite{fu2024topologic}                    & NeurIPS2024                        & \underline{25.9}   & \underline{}{54.7}   & \underline{21.6}    & \underline{17.9}    & \underline{42.3} \\
                            & TopoFormer(Ours)                     & -                               & {\bf 34.8(\textcolor{red}{+8.9})}      & {\bf 58.9(\textcolor{red}{+3.9})}      & {\bf 23.2(\textcolor{red}{+1.6})}       &{\bf 23.3(\textcolor{red}{+5.4})}        & {\bf 47.5(\textcolor{red}{+5.2})}    \\ \noalign{\smallskip} \hline\thickhline
\end{tabular}
\caption{{\bf Comparisons of our model and existing state-of-the-art methods on  $subset_A$ and  $subset_B$}~\cite{wang2023openlanev2}. ``-'' denotes the absence of relevant data. The best performances are highlighted in {\bf bold}, while the second one is \underline{underlined}. \textcolor{red}{Red} indicates the absolute improvements compared with the second one.}
\label{table: Traffic topology reasoning}

\end{table*}

\subsection{Datasets and Evaluation Setting}
\noindent{\bf OpenlaneV2 Dataset.}
The OpenLane-V2 dataset~\cite{wang2023openlanev2} presents two unique subsets, $subset_A$ and $subset_B$, which are derived from the Argoverse 2~\cite{wilson2023argoverse} and nuScenes~\cite{caesar2020nuscenes} datasets, respectively. Each subset comprises 1,000 scenes. For the $\text{T}^2\text{SG}$ generation task, we constructed a corresponding dataset based on OpenLane-V2, which contains 10 categories~\footnote{$\mathcal{C}_{lc}$= \{lane, go\_straight, turn\_left, turn\_right, no\_left\_turn, no\_right\_turn, u\_turn, no\_u\_turn, slight\_left, slight\_right\}} of the lane ({\it i.e.}, $\mathcal{C}_{lc}=10$). For the traffic topology reasoning task, we follow the same experimental setting in ~\cite{wang2023openlanev2}. We conducted the Traffic Topology Reasoning Task training based on $\text{T}^2\text{SG}$ and used the same MLP as in~\cite{li2023graph} to independently learn the bipartite graph between traffic lights and lanes.

\noindent{\bf Metric.} 
For the $\text{T}^2\text{SG}$ generation task, we adopt the Scene Graph Detection (SGDet) evaluation settings~\cite{im2024egtr} and report the Average Precision~(AP) of lane centerline detection which is class agnostic and mean Average Precision~(mAP) that aggregates the AP for each category. Following~\cite{wutopomlp}, the detection employs the Fréchet distance for quantifying similarity and we report the AP and mAP for the match thresholds set at \{1.0, 2.0, 3.0\}. For the edge in the scene graph, we compute the accuracy (A@1) as the evaluation metric. For the topology reasoning task, we utilize the OpenLane-V2~\cite{wang2023openlanev2} topology evaluation settings and report $\text{DET}_{l}$, $\text{DET}_{t}$, $\text{TOP}_{ll}$ and $\text{TOP}_{lt}$, which are the mAP on lane centerlines, traffic elements, topology among lanes and topology between lanes and traffic elements, respectively. To summarize the overall effect of primary detection and topology reasoning, the OpenLane-V2 Score (OLS) is calculated as follows:
\begin{equation}
    \text{OLS} = \frac{1}{4} \bigg[ \text{DET}_{l} + \text{DET}_{t} + f(\text{TOP}_{ll}) + f(\text{TOP}_{lt}) \bigg],
\end{equation}
where $f$ is the square root function.

\begin{figure*}[!t]
    \centering
    \includegraphics[width=0.9\linewidth]{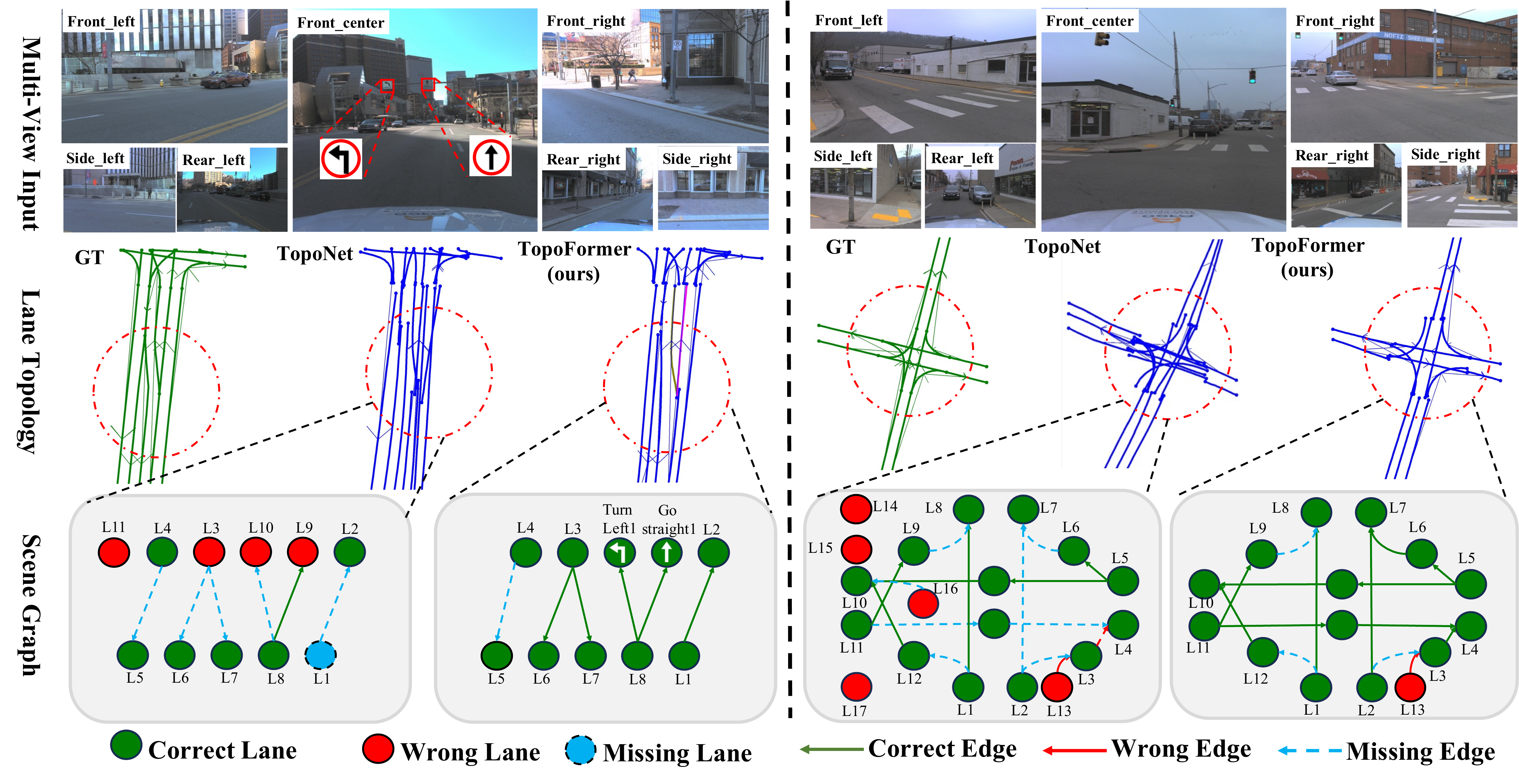}
    \caption{Qualitative results of the $\text{T}^2\text{SG}$ generation task and the lane topology reasoning, comparing the performance of TopoNet~\cite{li2023graph} and our proposed TopoFormer. The first row represents multi-view inputs. The second row illustrates the results of lane detection and lane topology reasoning. The third row visualizes our defined \( \text{T}^2\text{SG} \), with TopoNet's results converted to the same format for comparison. In these visualizations, green signifies correct predictions, red denotes erroneous predictions, and blue indicates missing predictions.}
    \label{fig: qualitativel}
\end{figure*}

\subsection{Implementation Details}
We implement our model based on Pytorch and MMdetection on 16 Tesla V100 GPUs with a total batch size of 16. All images are resized into the same resolution of 1550 × 2048, and we use ResNet-50~\cite{he2016deep} backbone pre-trained on ImageNet paired with a Feature Pyramid Network (FPN)~\cite{lin2017feature} to extract multi-scale features. The dimension of hidden feature $d$ is set to 256. The size of BEV grids is set to 200×100. For Lane centerline detection, the number of queries $N$ is 200, and the number of points in centerlines $l$ is 11. For the traffic elements detection, the number of queries is 100. The overall model TopoFormer is trained by AdamW optimizer with a weight decay of 0.01. The learning rate is initialized with 2e-4 and we employ a cosine annealing schedule~\cite{loshchilov2016sgdr} for the learning rate. 
Following~\cite{im2024egtr}, to accelerate convergence, we first train the lane detector and subsequently train the $\text{T}^2\text{SG}$ task using the pre-trained lane detector.

\subsection{Results}
\noindent{\bf $\text{T}^2\text{SG}$ generation.}
We present the quantitative performance of our proposed TopoFormer in \(\text{T}^2\text{SG}\) generation, comparing it with state-of-the-art scene graph generation methods in Table~\ref{table: T2SG}. The baseline reflects unprocessed input queries, while other methods focus on the semantic information related to traffic rules represented by the objects. Our approach emphasizes both semantic information and global lane dependencies, resulting in superior performance, In contrast, GCN-based methods like 3DSSG~\cite{wald2019rio} and EdgeGCN~\cite{zhang2021exploiting} exhibit low accuracy in edge prediction due to global modeling constraints, while our method aggregates global lane features using the geometric information of centerlines, leading to improved results. Specifically, in terms of edge accuracy, GCN-based methods (e.g., 3DSSG) significantly lag behind Transformer-based methods (e.g., SGformer) with scores of 8.5 vs. 0.4 in ${\text{A@1}}_{1.0}$ and 34.6 vs. 4.7 in ${\text{A@1}}_{3.0}$. Furthermore, when simultaneously modeling semantic information and global dependencies, our proposed TopoFormer outperforms Transformer-based methods in both node and edge accuracy, demonstrating its effectiveness in traffic scene graph generation tasks. 

\noindent{\bf Traffic topology reasoning.}
We present the quantitative performance of our TopoFormer in traffic topology reasoning in Table~\ref{table: Traffic topology reasoning}, our method surpasses other methods with a 46.3\% OLS in $subset_A$. Compared with TopoNet~\cite{li2023graph}, a graph-based method, our method achieved higher score in the topology reasoning task (24.1~{\it v.s.}~10.9 on $\text{TOP}_{ll}$, 29.5~{\it v.s.}~23.8 on $\text{TOP}_{lt}$) while also achieves decent centerlines detection score (34.7~{\it v.s.}~28.5 on $\text{DET}_{l}$).  This is attributed to $\text{T}^2\text{SG}$'s ability to capture global lane information via LAL, enhancing topology reasoning performance. Compared to methods like TopoLogic~\cite{fu2024topologic}, which also use geometric information, our TopoFormer improves $\text{DET}_{l}$ and $\text{TOP}_{lt}$ by effectively modeling road structure with CIL. Additionally, TopoFormer outperforms all models on $subset_B$, where centerlines are annotated in 2D space, demonstrating superior generalization.

\subsection{Ablation Studies}
\noindent{\bf Effects of the type of lane aggregation in LAL.}
To investigate the effectiveness of Geometry-guided Self-Attention, we introduced a variant labeled ``w/o \( A_{SPM} \)'', which excludes geometric information, along with three additional variations: ``Add'', ``Mul'', and ``Had'' representing the addition, multiplication, and Hadamard product of \( A_{SPM} \) with the self-attention weights, respectively. As shown in Table~\ref{table_ablation_CIM}, we focus on the performance of these variants in the lane-lane relationship~({\it i.e.}, {TOP$_{ll}$}). The incorporation of \( A_{SPM} \) leads to superior performance across all variants compared to ``w/o \( A_{SPM} \)'', highlighting the importance of geometric information. Among all methods, the Add variant demonstrates the highest performance.

\noindent{\bf Effects of the type of counterfactual intervention in CIL.} To investigate the effectiveness of counterfactual intervention, we introduced a variant labeled ``w/o CIL'', which excludes counterfactual intervention, along with three additional variations: ``CIL-Zero'', ``CIL-Mean'', ``CIL-Random'' representing the counterfactual interventions corresponding to zeros, the mean of attention weights, and randomly generated matrices, respectively. As shown in Table~\ref{table_ablation_CIM}, all implementations yield significant improvements over the ``w/o CIL'', demonstrating the generality of the proposed CIL. Furthermore, 1) the ``CIL-Zero'' exhibits marginally superior performance compared to the others, potentially because the zero matrices represent a completely untrue road structure, while the mean and random matrices still contain some possible structures, which better accentuates the causal impact of reasonable road structure on topological relationships. 2) The performance of these implementations is closely matched, indicating the robustness of the counterfactual intervention implementation, which can adapt to various implementations.

\begin{table}[]
\setlength{\tabcolsep}{5pt}
\renewcommand{\arraystretch}{1}
\begin{tabular}{lccccc}
\hline\thickhline
\multicolumn{1}{l}{{Methods}} & \multicolumn{1}{c}{{DET$_l$}} & \multicolumn{1}{c}{{DET$_t$}} & \multicolumn{1}{c}{{\cellcolor{gray!30}TOP$_{ll}$}} & \multicolumn{1}{c}{{TOP$_{lt}$}} & \multicolumn{1}{c}{{OLS}} \\ \hline
w/o $A_{SPM}$                 & 34.1                       & \underline{47.3}                       & \cellcolor{gray!30}21.6                        & 29.1                        & 45.4                    \\ 
w/ Had $A_{SPM}$             &  \underline{34.6}                      &   46.4                     & \cellcolor{gray!30}22.0                        & \underline{29.3}                        &     45.5                \\
w/ Mul $A_{SPM}$                &    {\bf 34.7}                    &   46.3                     & \cellcolor{gray!30}\underline{22.9}                        &  29.0                      &    \underline{45.7}                 \\
w/ Add $A_{SPM}$                & {\bf 34.7}                    & {\bf 48.2}                    & \cellcolor{gray!30}{\bf 24.1}                     & {\bf 29.5}                     & {\bf 46.3}                \\ \hline
w/o CIL                  & 32.2                       & 47.0                       &\cellcolor{gray!30}22.2                        & 28.6                        &44.9                     \\ 
CIL-Mean               & \underline{34.1}                       & \underline{47.5}                       & \cellcolor{gray!30}21.1                        & 28.4                        & 45.2                    \\
CIL-Random             & 34.0                       & 47.2                       & \cellcolor{gray!30}\underline{22.9}                        & \underline{28.8}                        & \underline{45.6}                    \\
CIL-Zero               & {\bf 34.7}                    & {\bf 48.2}                    & \cellcolor{gray!30}{\bf 24.1}                     & {\bf 29.5}                     & {\bf 46.3}                 \\ \hline\thickhline
\end{tabular}
\caption{Results of our TopoFormer with different variants on the traffic topology reasoning in OpenLane-V2 $subset_A$ set~\cite{wang2023openlanev2}. The best performances are highlighted in bold, while the second one is underlined. The gray shading part indicates we are more focused on TOP$_{ll}$.}
\label{table_ablation_CIM}
\end{table}

\subsection{Qualitative Analysis}
Figure~\ref{fig: qualitativel} presents a qualitative comparison between TopoNet~\cite{li2023graph} and our TopoFormer. The first row shows multi-view inputs of realistic scenes, while the second row displays lane topology results in the bird's eye view for both methods alongside the ground truth. The third row visualizes our defined \( \text{T}^2\text{SG} \), with TopoNet's results converted to the same format for comparison.
The results indicate that TopoFormer outperforms TopoNet in lane centerline detection and topology reasoning, accurately predicting most centerlines in both road structures ~({\it i.e.}, straight and intersections). Additionally, our \( \text{T}^2\text{SG} \) not only captures the traffic topology structure but also categorizes each lane, effectively demonstrating the learning of road signal semantics. 

%% file: sec/5_conclusion.tex
\section{Conclusion}
In this paper, we introduced a new traffic topology scene graph (\( \text{T}^2\text{SG} \)) for traffic scene understanding and presented the TopoFormer, a one-stage topology scene graph transformer for \( \text{T}^2\text{SG} \) generation. TopoFormer features a Lane Aggregation Layer for global lane feature aggregation and a Counterfactual Intervention Layer to explore the road structure of the traffic scene. Our experiments demonstrate that TopoFormer outperforms state-of-the-art methods in \( \text{T}^2\text{SG} \) generation and significantly enhances traffic topology reasoning on the OpenLane-V2 benchmark.

%% file: sec/6_acknowledgments.tex
\section{Acknowledgement}
This work is partly supported by the Funds for the National Natural Science Foundation of China under Grant 62202063 and U24B20176,  Beijing Natural Science Foundation (L243027), Beijing Major Science and Technology Project under Contract No. Z231100007423014.